\pdfoutput=1

\documentclass[11pt]{article}
\usepackage[]{emnlp2022}

\usepackage{times}
\usepackage{latexsym}
\usepackage{amsmath}

\usepackage[T1]{fontenc}

\usepackage[utf8]{inputenc}

\usepackage{microtype}

\usepackage{graphicx} 
\usepackage{amsfonts}
\usepackage{arydshln}
\usepackage{multirow}
\usepackage{subcaption}
\usepackage{booktabs}

\newtheorem{proposition}{Proposition}

\title{Hard Gate Knowledge Distillation - \\Leverage Calibration for a Robust and Reliable Language Model} 
\author{Dongkyu Lee$^{1,3}$\Thanks{This work was done while Dongkyu was an intern at NVIDIA}\quad Zhiliang Tian$^{2}$\Thanks{Corresponding author}\quad Yingxiu Zhao$^{1}$ \\\quad \textbf{Ka Chun Cheung}$^{3}$ \quad \textbf{Nevin L. Zhang}$^{1}$\\$^{1}$Department of Computer Science and Engineering, HKUST \\$^{2}$College of Computer, National University of Defense Technology \\$^3$NVIDIA AI Technology Center, NVIDIA\\$^{1}$\texttt{\{dleear, yzhaocx, lzhang\}@cse.ust.hk}\\$^2$\texttt{tianzhilianghit@gmail.com}\quad $^{3}$\texttt{chcheung@nvidia.com}}

\begin{document}
\maketitle
\begin{abstract}
  In knowledge distillation, a student model is trained with supervisions from both knowledge from a teacher and observations drawn from a training data distribution. Knowledge of a teacher is considered a subject that holds inter-class relations which send a meaningful supervision to a student; hence, much effort has been put to find such knowledge to be distilled. In this paper, we explore a question that has been given little attention: ``\emph{when to distill such knowledge}." The question is answered in our work with the concept of model calibration; we view a teacher model not only as a source of knowledge but also as a gauge to detect miscalibration of a student. This simple and yet novel view leads to a hard gate knowledge distillation scheme that switches between learning from a teacher model and training data. We verify the gating mechanism in the context of natural language generation at both the token-level and the sentence-level. Empirical comparisons with strong baselines show that hard gate knowledge distillation not only improves model generalization, but also significantly lowers model calibration error.
\end{abstract}
\section{Introduction}
In recent years, the deep learning community has achieved marked performance gains across a variety of tasks \citep{gpt3,devlin2018pretraining}. In the meantime, some deep learning models have become excessively large, limiting their applicability in some scenarios. To cope with the issue, \citet{KD} proposed knowledge distillation (KD), in which knowledge of a large network, called a teacher network, is transferred to a relatively small model, called a student model. 

The benefits of KD have been widely witnessed across multiple domains \citep{FitNets, tinyBert}. Recently, it has been observed that KD can be used in both reducing model size and improving model generalization \citep{understanding, pmlr-v80-furlanello18a}. \citet{KD} argue that a distribution, defined by a teacher, holds inter-class relations, commonly referred to as the \emph{dark knowledge}, and that such distribution brings a meaningful supervision to a student. Therefore, a large body of research in KD has viewed a teacher as a source of knowledge and has focused on \emph{finding a meaningful knowledge} to be transferred \citep{FitNets,dropoutDistillation, rkd,tf-kd,ps-kd}.

In this work, we focus on \emph{when to distill knowledge of a teacher}. 
This is a central question to ask, as a model can benefit from the adaptive control of supervision between ground truth and a teacher; 
When a model is trained to increase the predictive score of a prediction, a one-hot encoded supervision, without incorporating teacher model, sends a direct signal in increasing the score \citep{whenDoes}.
In another case, when a model is trained to learn knowledge of a teacher, a teacher's output without fusing a ground truth sends more direct signal in minimizing the knowledge gap between the student and the teacher. 
However, the question of ``when" has not been answered. For this reason, previous works choose to learn from both of the supervisions. 

We give an answer to the question from the perspective of model calibration. Model calibration refers to how well a predicted probability of a model reflects the true accuracy. Therefore, a well-calibrated predictive score represents the \textbf{likelihood of correctness of a prediction} \citep{pmlr-v70-guo17a}. In this light, such score can be viewed as a gauge to detect a miscalibration of a student in training; when a student makes a prediction with a probability mass that is higher than the expected accuracy of the prediction (overconfidence), a student model is trained with only supervision from a teacher. In the case of underconfidence, a student is trained with only supervision from ground-truth. 

Switching supervision is supported by two widely accepted ideas: 1) the close link between miscalibration and overfitting, and 2) the regularization effect of KD. \citet{pmlr-v70-guo17a} empirically find that a model overfits to negative log likelihood (NLL) training, leading to miscalibration, and \citet{NEURIPS2020_aeb7b30e} further support the claim. Therefore, we utilize the regularization effect held in KD training \citep{tf-kd}. Aside from the inter-class relations held in knowledge, recent findings suggest that KD is a form of adaptive regularization \citep{understanding, tf-kd}, where a teacher enforces a student to distribute probability mass on output space more evenly.

Taking all these factors into account, we present a simple, yet novel KD method, called Hard gate Knowledge Distillation (HKD). Given a calibrated teacher model, the teacher gates supervisions between knowledge and observation for each instance/time step, selecting which objective the student should be optimized to. We introduce two levels of hard gates: the token-level and the sentence-level which are instance-specific hard gates computed on the fly during forward propagation. Our work validates the proposed idea on a task in the Natural Language Generation (NLG) domain, as there is an inseparable relation between the quality of an output and model calibration \citep{DBLP:journals/corr/abs-1903-00802}. 

The contributions of the proposed method are as follows:
\begin{itemize}
  \item To the best of our knowledge, this work is the first attempt to leverage knowledge and observations in KD with a hard gate which is instance-specific.
  \item Our work introduces a novel view and role of a teacher model in student-teacher framework which improve model generalization and model calibration of a student by a significant margin across multiple datasets.
\end{itemize}

\section{Preliminaries \& Related Work}
\subsection{Knowledge Distillation}
The conventional logit-based KD \citep{KD} aims to minimize the distance between the probability distribution mapped by a teacher and that of a student, while another objective is to maximize the likelihood of predicting ground truth. Following is the loss of an instance $(\mathbf{x}^i, \mathbf{y}^i) \in \mathcal{X}\times \mathcal{Y}$ at time-step $t$, where $i$ indicates the index of the sample.\footnote{Loss equations are illustrated in time-step level hereinafter as a natural language generation task can be viewed as a sequence of classification.}
\begin{equation}\label{eq:kd}
    \begin{split}
  \mathcal{L}_{kd} &= -\sum_v^{|V|}(1-\alpha) y_{t,v}^{i}\log P_\theta(y_{t,v}^{i}|\mathbf{c}_{<t}^{i})\\
  &+\alpha P_\phi(y_{t,v}^{i}|\mathbf{c}^i_{<t};\tau)\log P_\theta(y_{t,v}^i|\mathbf{c}_{<t}^i;\tau)\\
  \end{split}
\end{equation}
$V$ and $\tau$ denote a set of vocabularies and a temperature respectively. $\phi$ and $\theta$ denote parameters of a teacher and those of a student. $\alpha$ denotes a balancing parameter which in this work is termed a \textbf{gate}, and $\mathbf{c}_{<t}$ is a context at time step $t$, hence made of input $\mathbf{x}$ and preceding tokens $\mathbf{y}_{<t}$. The gate is set to a value between 0 and 1, which indicates a \emph{soft gate, and it is shared among instances and remains fixed throughout training} \citep{rkd, KD, tf-kd}. Therefore, a student model is trained with a soft target $\tilde{y}^i_t$, a result of linear interpolation between a ground truth and a distribution mapped by a teacher. 

Numerous studies have attempted to find meaningful knowledge to be distilled. Starting with inter-class relations on logit space \citep{rkd, KD}, the scope of knowledge expanded to feature-level \citep{FitNets} to encourage a student to maintain similar intermediate representations to those of a teacher. Recent studies find that even a model with an identical model structure to that of a student can suit the role as a teacher; thus it is commonly referred to as Self-Knowledge Distillation \citep{tf-kd,ps-kd,liu-etal-2021-noisy}. \citep{tf-kd,understanding} argue that the success is brought by KD's close link to label smoothing \citep{label_smoothing}, with KD holding a regularization effect. In this regard, there have been attempts to explore the importance of the soft gate. PS-KD \citep{ps-kd} linearly increases the value of the gate in the course of training. Similar to our work, \citet{customizedKD} propose a hard gate mechanism in KD; however the work utilizes an iteration-specific hard gate, and the gates only apply to distillation loss of KD. 
\subsection{Calibration}
A model is said to be well-calibrated when the predictive confidence truly reflects true accuracy \citep{pmlr-v70-guo17a}.
\begin{equation}
  P(\hat{Y}=Y|P(\hat{Y}|X)=p) = p, \forall p \in [0,1]
\end{equation}
Therefore, when a model makes predictions with probability of $p$, the accuracy of the predictions is expected to be $p$. The quantity is commonly approximated with Expected Calibration Error and Maximum Calibration Error \citep{ece}.

There have been continuous efforts in lowering the calibration error of a model, and one of the simplest, yet effective methods is temperature scaling \citep{pmlr-v70-guo17a}. Temperature scaling is a parametric post-hoc calibration method, where a single parameter is learned; with model parameters fixed, the single parameter is learned to lower the negative log likelihood on validation dataset. This simple calibration method has been widely appreciated for its ability to improve the reliability of a model \citep{whenDoes}.

\section{Approach}
\begin{figure*}[t]
  \centering
  \includegraphics[width=0.9\textwidth]{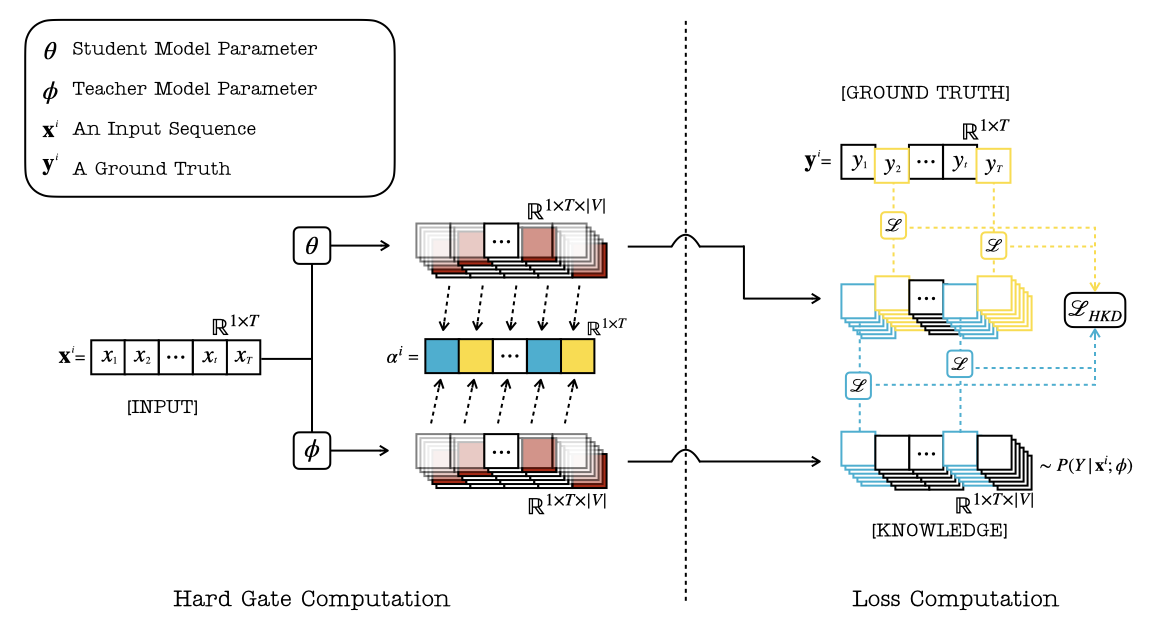}
  \caption{Overview of the proposed token-level hard gate KD. The conditional probability distributions mapped by a student and a teacher are in comparison to compute the instance and time step-specific hard gates. The loss of the instance is computed according to the hard gates.}
  \label{modelStructure}
  \end{figure*}
In this section, we first discuss a new interpretation of a teacher under KD training and introduce methods that switch supervision between knowledge and observations with an instance-specific hard gate.
\subsection{A View on Teacher Model}\label{sec:selfTeacher}
When a teacher model is well-calibrated, via calibration method such as temperature scaling \citep{pmlr-v70-guo17a}, the predictive score of a teacher can be used to \emph{estimate} the true likelihood of correctness. In this light, a teacher can be used to evaluate if a student model makes a miscalibrated prediction, either resulting in underconfidence or overconfidence. Furthermore, given a calibrated teacher, minimizing the knowledge gap provides a meaningful insight which is more than learning the inter-class relations, as the objective \emph{extends to improving calibration} of a student. By minimizing the KL divergence between the two probability distributions, the prediction of a student is expected to reflect the calibrated predictive score. 


\subsection{Hard Gate}
From the novel view of a teacher, our work presents two instance-specific hard gates: the token-level and the sentence-level hard gate. 
\subsubsection{Token-Level Gate}
When a predictive score of a prediction by a student is high compared to an \emph{approximated} likelihood of the correctness of the prediction, a student is supervised to distribute the probability mass to other remaining classes, hence learning to output a smooth distribution. In another case, in which the predictive score is less than the approximation, the student is learned with supervision that increases the probability, learning from a sample drawn from the data distribution.

In every time step, instance-specific hard gates are computed on the fly during forward propagation as follows:
\begin{equation}\label{eq:indicator1}
  a^{i}_t = 
\begin{cases}
    1,& \text{if } P_{\theta}(y^i_{t, j}|\mathbf{c}^i_{<t})> f(y^i_{t, j}, \mathbf{c}^i_{<t}) \\
    0,& \text{otherwise}
\end{cases}
\end{equation}
$P_\theta(y^i_{t, j}|\mathbf{c}^i_{<t})$ and $f(y^i_{t,j}, \mathbf{c}^i_{<t})$ are conditional probability of a ground truth index $j$ mapped by a student model and the true likelihood of $y^i_{t,j}$ occurring in the given context. Since the true likelihood of correctness cannot be obtained, we approximate the quantity with a teacher network with enhanced calibration ability $f(y^i_{t,j}, \mathbf{c}^i_{<t}) \approx P_\phi(y^i_{t,j}|\mathbf{c}_{<t};\tau)$. 

\paragraph{Supervision from Observations ($\alpha=0$)}
When the hard gate is computed to be 0, it is an indication of \emph{underfitting and underconfidence} by a student on the instance. The student needs further training so that the likelihood of predicting the target index is increased. Due to the normalizing activation layer, softmax, a direct way of escalating the probability mass on the ground truth index is to minimize the KL divergence with one-hot encoded ground truth \citep{whenDoes}, without incorporating knowledge. That being the case, when the hard gate is set to 0, supervision to a student solely comes from ground truth.
\paragraph{Supervision from Knowledge ($\alpha=1$)}
In another case when the gate is set to 1, it is an indication of \emph{overconfidence} evaluated by the approximated quantity mapped by a teacher. Therefore, a student is trained to distribute the probability mass on output space more evenly; the student learns to close the gap between its probability distribution and that of a teacher. 

This gating mechanism can be viewed as smoothing of labels, hence presenting a regularization effect. 
The entropy of supervisions by the proposed method, conventional logit-based KD ($\tilde{\mathbf{y}}^i_t$), and one-hot encoded target (hard target) are as follows:
\begin{equation}
  H(P_\phi(Y|\mathbf{c}^i_{<t};\tau))\geq H(\tilde{\mathbf{y}}^i_t)\geq H(\mathbf{y}^i_t)
\end{equation}
where $\tilde{\mathbf{y}}^i_t$ is the soft target that is a linear interpolation of a ground truth and a probability distribution mapped by a teacher. The entropies illustrate how the proposed method regularizes a student by presenting high entropy supervision. Specifically, the following proposition holds.

\begin{proposition}[Opposite Gradient]\label{pro:oppositeGradient}
  When $\alpha = 1$, the sign of the expectation of the gradient by the proposed KD method with respect to logit on ``incorrect" classes is guaranteed to be opposite to that of the cross entropy with hard target.
\begin{equation}
  \mathbb{E}_{i\neq j}\frac{\partial\mathcal{L}_{hkd}}{\partial z_i}<0\leq\mathbb{E}_{i\neq j}\frac{\partial\mathcal{L}_{ce}}{\partial z_i}
\end{equation}
\end{proposition}
The gradient of a sample with respect to a logit $z$ by the cross entropy is as follows\footnote{For notational simplicity, time step is omitted.}:
\begin{equation}
  \frac{\partial \mathcal{L}_{ce}}{\partial z_i} = P_{\theta}(y_i|\mathbf{c}_{<t})-y_i
\end{equation}
When $\alpha =1$, the gradient of the proposed method is
\begin{equation}
  \frac{\partial \mathcal{L}_{hkd}}{\partial z_i} = P_{\theta}(y_i|\mathbf{c}_{<t})-P_{\phi}(y_i|\mathbf{c}_{<t};\tau)
\end{equation}
Then, it is straightforward to compute the sum of the quantities except the target index $j$.
\begin{equation}\label{eq:ce_incorrect}
  \sum_{i, i\neq j}\frac{\partial \mathcal{L}_{ce}}{\partial z_i} = 1-P_{\theta}(y_j|\mathbf{c}_{<t})
\end{equation}
\begin{equation}\label{eq:hkd_incorrect}
  \begin{aligned}
  \sum_{i, i\neq j}\frac{\partial \mathcal{L}_{hkd}}{\partial z_i} &= P_\phi(y_j|\mathbf{c}_{<t}) - P_\theta(y_j|\mathbf{c}_{<t})
  \end{aligned}
\end{equation}
As Equation \ref{eq:ce_incorrect} is guaranteed to be greater than or equal to 0, Equation \ref{eq:hkd_incorrect} must be smaller than 0, since $P_\phi(y_j|\mathbf{c}_{<t}) <  P_\theta(y_j|\mathbf{c}_{<t})$. The Proposition \ref{pro:oppositeGradient} is not guaranteed in conventional logit-based KD, while it holds true within the proposed approach.

The cross entropy with hard target forces a student to decrease the probability mass on the other classes, while the proposed method sends gradients that have opposite direction to that of the cross entropy. In return, the proposed method pushes a student to increase the probability mass on the other output space, regularizing the student. 

In addition to the regularization effect, the inter-class relations are given more directly to the student than that of the conventional logit-based KD. The conventional KD shrinks the dark knowledge $P_\phi(y_{t,v}^{i}|\mathbf{c}^i_{<t};\tau)$ by $\alpha$ as in Equation \ref{eq:kd}. This, however, is different in the proposed method as $\alpha$ is set to 1, and hence the amount of dark knowledge remains unchanged.

\subsubsection{Sentence-Level Gate}
A natural language generation task is a sequence of classification. In this regard, in addition to the token-level gate, we propose to compute hard gates on the sentence-level. In particular, the gates are determined by comparing the sentence probabilities mapped by the two models in KD. 
\begin{equation}\label{eq:sentence}
  \forall_t a^{i}_t = 
\begin{cases}
    1,& \text{if } P_\theta(\mathbf{y}^i|\mathbf{x}^i)> f(\mathbf{y}^i, \mathbf{x}^i)\\
    0,              & \text{otherwise}
\end{cases}
\end{equation}
where $P(\mathbf{y}^i|\mathbf{x}^i)$ is a sentence probability computed as the product of the conditional probabilities of time steps $\prod P_\theta(y^i_{t,j}|\mathbf{c}^i_{<t})$. 
As in the token-level gate, the true likelihood of the sentence appearing is approximated with a teacher $f(\mathbf{y}^i, \mathbf{x}^i) \approx \prod_t P_\phi(y^i_{t,j}|\mathbf{c}^i_{<t}) $.

A probability of a sentence defined by a language model is a reflection of how likely a model predicts the sentence. If a student model defines a sentence probability that is higher than that of a calibrated teacher, this is a possible sign of overconfidence in the sentence. Therefore, in such case, the student only receives supervision from knowledge. In the opposite case, as in the token-level gate, the student is solely trained with ground truth. 

Sentence-level computes hard gates in a more cautious manner than token-level does. A sentence probability is a product of probabilities of words within the sentence; hence a miscalibrated probability of a word can cause much change in the final probability. 
This aspect is depicted in Figure \ref{fig:ExpectedAlpha} in which sentence-level gates and token-level gates differ in the ratio of $\alpha$ in the course of training.

\subsection{Final Loss}
\begin{figure}[t]
    \centering
    \includegraphics[width=0.8\columnwidth]{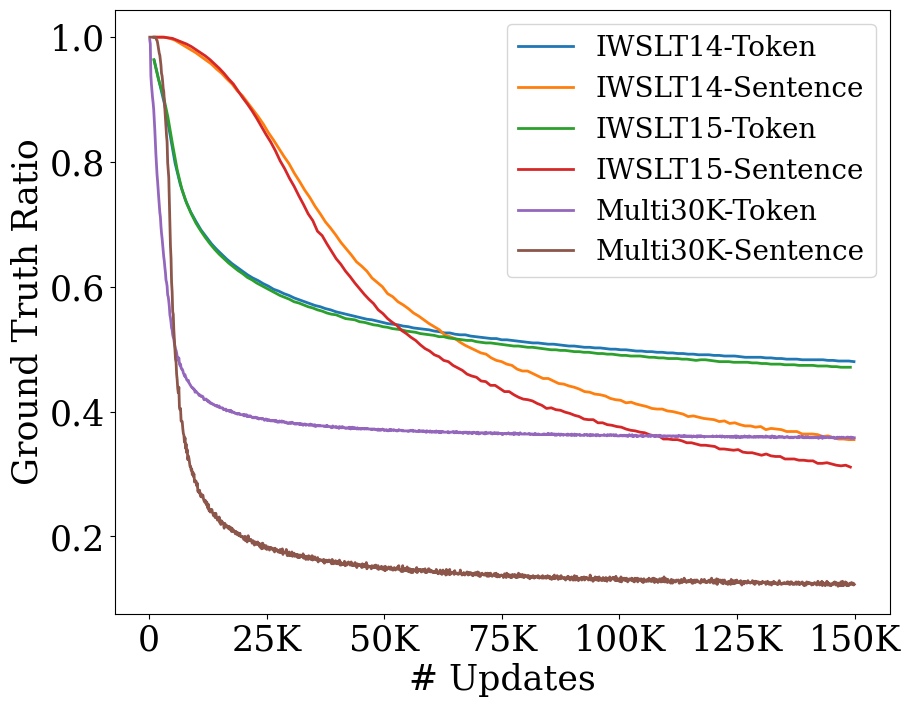}
    \caption{This figure plots the change of ground truth supervision ratio in the course of training.}
    \label{fig:ExpectedAlpha}
    \end{figure}

The final loss function is as follows:
\begin{equation}\label{eq:hkdLoss}
    \begin{split}
  \mathcal{L}_{hkd} &= -\sum_v^{|V|}(1-\alpha^i_t) y_{t,v}^{i}\log P(y_{t,v}^{i}|\mathbf{x}^{i};\theta)\\
  &+\alpha^i_t P_\tau(y_{t,v}^{i}|\mathbf{x}^i;\phi)\log P(y_{t,v}^i|\mathbf{x}^i;\theta)
  \end{split}
\end{equation}
The $\alpha$ in both token and sentence-level is computed during the forward propagation. Therefore, the following propositions can be made.
\begin{proposition}{When expected $\alpha$ approaches 0, $\mathcal{L}_{hkd}$ reduces to MLE with hard targets. In other case, when the expected value approaches 1, $\mathcal{L}_{hkd}$ reduces to minimizing KL divergence between probability distribution of a student and that of a teacher.}
\end{proposition}
\begin{equation}\label{eq:lim1}
  \begin{aligned}
  \lim_{\mathbb{E}[\alpha] \to 0}\mathcal{L}_{hkd} &= \mathcal{L}_{ce}(P(Y|X;\theta),Y)\\
  \lim_{\mathbb{E}[\alpha] \to 1}\mathcal{L}_{hkd} &= \mathcal{L}_{ce}(P(Y|X;\theta), P_\tau(Y|X;\phi))
  \end{aligned}
\end{equation}
where $\mathcal{L}_{ce}$ indicates the cross-entropy loss.
We empirically observe that the former case is seen in the early stages of training which is depicted in Figure \ref{fig:ExpectedAlpha}. In the other case in which the expected value converges to 1, the loss reduces to solely minimizing the distance of the distributions mapped by the models without any observation from empirical training distribution.


One difference to notice is the temperature scaling in Equation \ref{eq:hkdLoss}. The proposed KD solely applies temperature scaling on the logit of a teacher for the purpose of calibrating the teacher's output. This is a marked difference from the existing logit-based KD, where both a student and a teacher logits are scaled with a pre-defined temperature as in Equation \ref{eq:kd}. This distinction encourages a student model to mimic a probability distribution of a teacher which contains inter-class relations as well as calibrated predictive scores.
\begin{table*}[t]
  \centering
  \resizebox{0.95\textwidth}{!}{
  \begin{tabular}{cccc:cccccc}
  \hline
  &&\multicolumn{2}{c}{\textbf{Calibration}}&\multicolumn{5}{c}{\textbf{Generalization}}\\
  \cmidrule{3-4} \cmidrule{5-9}
  \textbf{Dataset}&\textbf{Method}&\textbf{ECE} ($\downarrow$)&\textbf{MCE} ($\downarrow$) &\textbf{BLEU} &\textbf{METEOR} & \textbf{WER} ($\downarrow$)&\textbf{ROUGE-L} & \textbf{NIST} \\
  \hline
  \multirow{11}{*}{\texttt{Multi30K}}&Base&14.95&26.01&40.64 &	73.31 &	38.76	&69.21&7.96\\
  \cdashline{2-9}
  &LS-Uniform&9.17&17.22&42.55&74.78&38.00&70.17&8.08\\
  &LS-Unigram&9.12&17.78&42.52&74.61&37.54&70.30&8.13\\
  &ConfPen&48.21&52.53&43.14&75.03&37.95&70.54&8.12\\
  &Loras&20.27 &40.86&41.78&74.31&38.03&69.83&8.12\\
  \cdashline{2-9}
  &TF-KD&21.18&42.87&41.77&74.33&37.55&69.94&8.12\\
  &PS-KD&14.75&26.69&41.95&74.26&38.45&69.75&8.05\\
  &SD&6.87 & 12.38 &43.41&75.41&37.50&70.76&8.15\\
  &Beta&11.71 & 21.90&41.9&74.24&38.56&69.66&8.03\\
  \cdashline{2-9}
  &HKD-T (Ours)&\underline{2.86}&\underline{6.92}&\textbf{43.96}&\textbf{75.64}&\textbf{36.38}&\textbf{71.32}&\underline{8.26}\\
  &HKD-S (Ours)&\textbf{2.25}&\textbf{3.78}&\underline{43.78}&\underline{75.48}&\underline{36.58}&\underline{71.20}&\textbf{8.27}\\
  \hline
  \hline
  \multirow{11}{*}{\texttt{IWSLT15}}&Base&14.05&20.38&30.17&59.09&54.02&63.91&7.18\\
  \cdashline{2-9}
  &LS-Uniform&8.53&12.13&30.74&59.7&53.62&64.33&7.24\\
&LS-Unigram&7.89&11.71&30.8&59.57&53.23&64.34&7.27\\
&ConfPen&43.94&59.28&31.10&59.65&53.00&64.50&7.31\\
&Loras&12.41 & 19.15&30.13&58.98&53.79&63.72&7.21\\
  \cdashline{2-9}
&TF-KD&13.30&19.29&30.17&58.94&54.01&63.83&7.16\\
&PS-KD&9.34&14.18&31.21&60.04&52.71&64.71&7.34\\
&SD&5.01 & 9.64&30.86&59.54&53.39&64.36&7.27\\
&Beta&9.57 & 14.97&30.48&59.29&53.39&64.07&7.26\\
  \cdashline{2-9}
  &HKD-T (Ours)&\underline{2.17}&\underline{5.12}&\textbf{31.96}&\textbf{60.54}&\textbf{52.21}&\textbf{65.01}&\textbf{7.40}\\
  &HKD-S (Ours)&\textbf{1.38}&\textbf{4.70}&\underline{31.85}&\underline{60.35}&\underline{52.24}&\underline{64.94}&\underline{7.39}\\
  \hline
  \hline
  \multirow{11}{*}{\texttt{IWSLT14}}&Base&12.98&19.29&35.96 &64.70 &48.17&61.82 &8.47\\
  \cdashline{2-9}
  &LS-Uniform&6.43&9.98&36.82&65.30&47.50&62.41&8.60\\
  &LS-Unigram&6.12&9.46&36.97&65.38&47.30&62.57&8.62\\
  &ConfPen&48.19&57.58&37.11&65.55&47.09&62.67&8.66\\
  &Loras&10.54 & 15.29&36.32&64.93&48.78&61.98&8.43\\
  \cdashline{2-9}
  &TF-KD&12.20&17.60&36.35&64.88&47.84&62.11&8.53\\
  &PS-KD&5.63&8.88&37.49&65.81&46.48&63.07&8.71\\
  &SD& 7.82 & 13.71&37.35&65.76&47.33&62.70&8.63\\
  &Beta&8.63 & 13.21&36.91&65.39&47.87&62.40&8.55\\
  \cdashline{2-9}
  &HKD-T (Ours)&\underline{1.43}&\underline{3.52}&\textbf{38.27}&\textbf{66.54}&\underline{45.74}&\textbf{63.73}&\textbf{8.85}\\
  &HKD-S (Ours)&\textbf{1.27}&\textbf{3.22}&\underline{38.08}&\underline{66.44}&\textbf{45.72}&\underline{63.65}&\underline{8.84}\\
  \hline
  \end{tabular}}
  \caption{Scores are reported in percentage by averaging three runs with different random seeds. A bold number indicates the best performance within each corpus tested, and the underlined numbers are the second best performing scores. HKD-T and HKD-S denote the proposed method with the token-level and the sentence-level hard gates respectively.}
  \label{table:evaluation1}
\end{table*}
\begin{figure*}[ht]
  \begin{subfigure}{.33\textwidth}
    \centering
    \includegraphics[width=\linewidth]{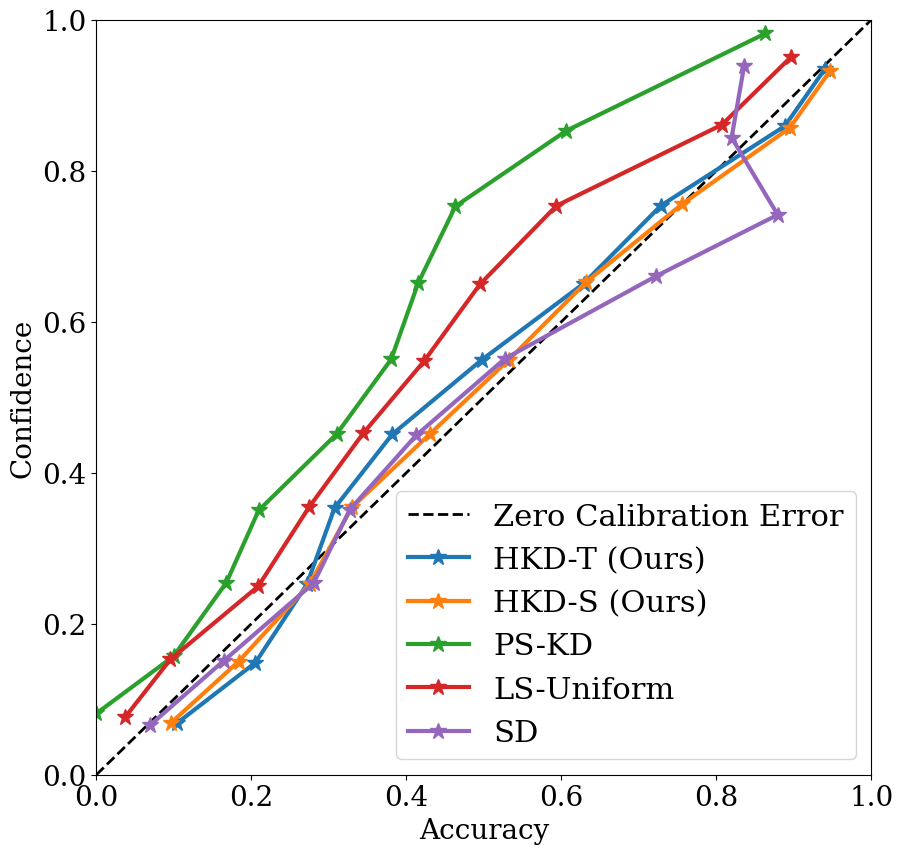}
    \caption{Multi30K DE-EN }
    \label{fig:rel-multi30k}
  \end{subfigure}
  \begin{subfigure}{.33\textwidth}
    \centering
    \includegraphics[width=\linewidth]{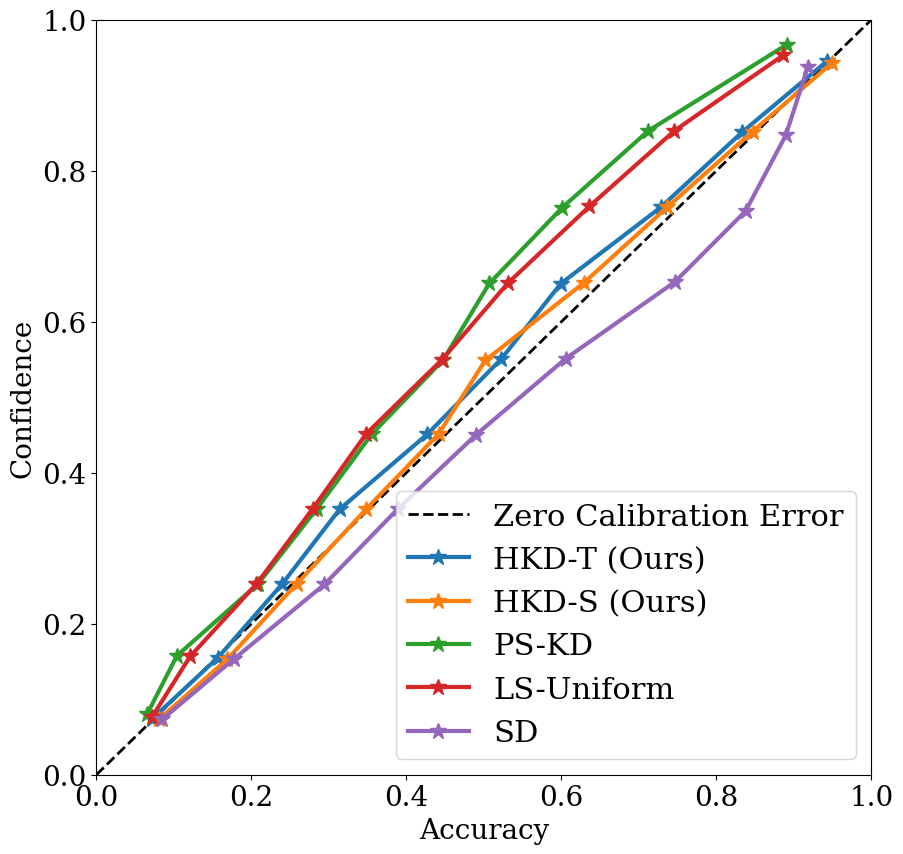}
    \caption{IWSLT15 EN-VI}
    \label{fig:rel-iwslt15}
  \end{subfigure}
  \begin{subfigure}{.33\textwidth}
    \centering
    \includegraphics[width=\linewidth]{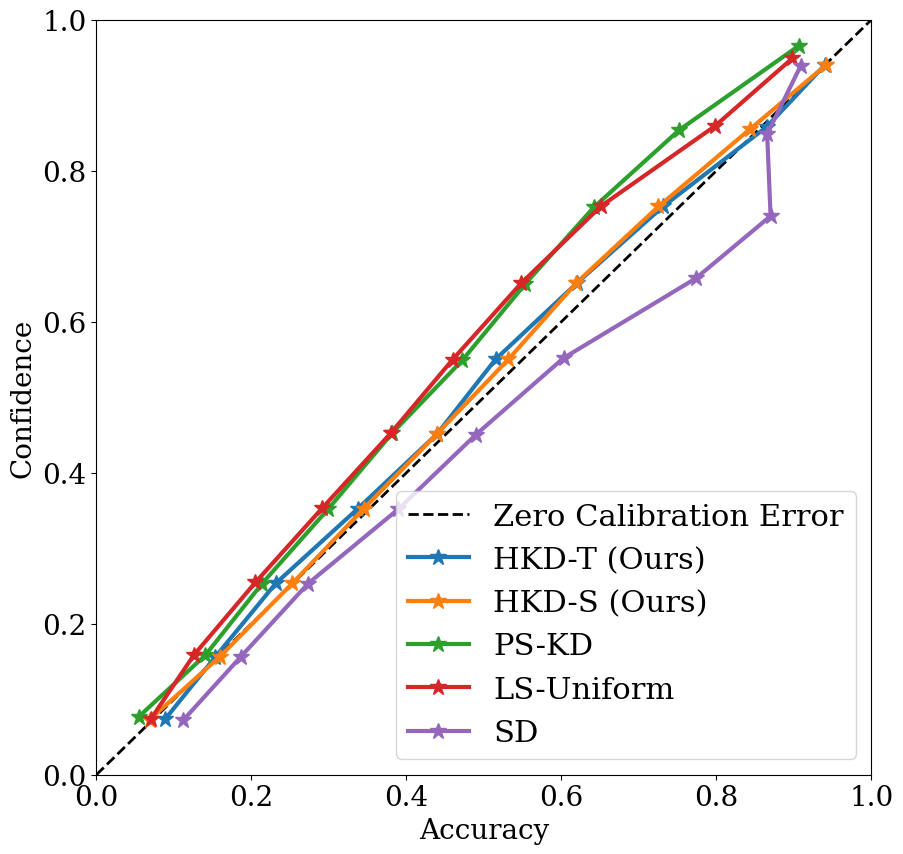}
    \caption{IWSLT14 DE-EN}
    \label{fig:rel-iwslt14}
  \end{subfigure}
  \caption{Reliability diagrams of the five methods on the three corpora tested. The dashed line indicates zero calibration error.}
  \label{fig:reliability}
  \end{figure*}

\section{Experiment} 
\subsection{Dataset \& Experiment Setup}
We validate the proposed gating mechanisms on three popular translation tasks: IWSLT14 German to English (DE-EN), IWSLT15 English to Vietnamese (EN-VI), and Multi30K DE-EN\footnote{The details of the datasets are reported in Appendix \ref{sec:Appen-data}.}. There are two core reasons for conducting experiments on a NLG task. First, our method suits NLG tasks by nature (token and sentence-level). Second, calibration has inseparable relation with NLG, as popular generation schemes, such as top-$k$, top-$p$, and beam search, are affected by the calibration ability of a language model. The generation schemes start by assuming that a predictive score represents likelihood of an event \citep{whenDoes}.

All of the experiments are conducted on a single Telsa V100, and both a student and a teacher model follow transformer architecture \citep{transformer}. The proposed method is tested on self-knowledge distillation environment, considering the efficiency of computation and general applicability\footnote{Self-knowledge distillation can be used even when strong teacher does not exist or cannot be obtained \citep{tf-kd}.}. The hyperparameters are identical to the specified configuration in \texttt{fairseq} \citep{fairseq}\footnote{\url{https://github.com/facebookresearch/fairseq/blob/main/examples/translation/README.md}}. The teacher network is trained with uniform label smoothing \citep{label_smoothing} in our environment; nonetheless a teacher trained with the regular cross-entropy training with hard targets is a valid option, which we show in the ablation study. 
For evaluation, we comprehensively validate previous methods and the proposed methods with popular translation evaluation metrics: BLEU \citep{bleu}, METEOR \citep{meteor}, Word Error Rate (WER), ROUGE-L \citep{rouge}, and NIST \citep{nist}. For quantifying the level of model calibration, we report Expected Calibration Error (ECE) and Maximum Calibration Error (MCE) \citep{ece}.
\subsection{Baselines}
As this paper lies in a branch of KD, though is closely linked to regularization methods, we compare the proposed methods with baselines from both of the domains.
\paragraph{Base Method} The base method in this work is the cross-entropy with hard targets (Base).
\paragraph{Regularizers} Although Label Smoothing (LS) was first introduced to enhance model performance in \citep{label_smoothing}, it has been found to help in model calibration as well. The prior label distribution is commonly set with a uniform distribution (LS-Uniform) \citep{transformer, bart}, yet unigram distribution (LS-Unigram) is another valid choice. Similar to label smoothing, ConfPen \citep{conf_penalty} prevents a model from outputting a peak distribution by penalizing high confident predictions. Loras \citep{loras} theoretically find that the generalization error largely depends on prior label distribution; thus, it jointly learns model parameters and prior label distribution. 
\paragraph{KD-Methods} \citet{tf-kd} empirically find that even a weak teacher can improve a student. Accordingly, the authors present TF-KD where a pre-trained teacher with identical model structure to that of a student is utilized in KD. PS-KD \citep{ps-kd} is similar, but the core difference is that a teacher is the previous checkpoint of a student in training. \citet{instanceSpecific} introduce instance-specific label smoothing methods, SD and Beta, which make use of self-knowledge and encourage diversity in predictions.

\subsection{Experimental Result}
The automatic evaluation results are reported in Table \ref{table:evaluation1}. Both of the proposed methods achieve noticeable gains compared to the Base method and the strong baseline methods. The improvements are seen across every metric and corpus tested. Our methods excel not just in $n$-gram matching (BLEU) and harmonic mean of unigram precision and recall (METEOR), but also in having the longest common subsequence with references (ROUGE-L) and outputting informative $n$-grams (NIST). Moreover, as clearly depicted with low WER, the outputs of our systems require the least amount of modification to be converted into reference sequences. Without adding any additional learnable parameter compared to the base method, our token-level HKD illustrates superior performance to that of the base method, absolute gain of 3.32 BLEU score and relative improvement of 8.16\% on Multi30K. Sentence-level hard gate method also illustrates competitive results to those of the token-level hard gate. On every corpus and metric tested, both token and sentence-level gates outperform the strong baselines by a large margin.

Figure \ref{fig:ExpectedAlpha} depicts the change of ground truth supervision ratio in the course of training. In each corpus, the ratio of ground truth decays throughout the training. In the early stage of training, most of the supervisions come from ground truth as a student is underfitted, leading to the low expected $\alpha$. The ratio decreases as the student model learns to map the task distribution, increase in the ratio of supervision from knowledge. The ratios converge at a certain point in each dataset, illustrating how the proposed methods hold self-regulating property in switching the supervisions. A noteworthy point is the \emph{correlation of the ratio and the corpus size}. Under Multi30K training, which is the smallest in terms of training dataset size, the student receives majority of supervision from the teacher which the ratio is higher than 0.6 with token-level and 0.8 with sentence-level hard gate. This empirically shows that the proposed systems, in an environment with the risk of overconfidence and overfitting demonstrate strong regularization effect in training. The methods enforce a student model to learn from knowledge to avoid possible degradation in model generalization and calibration.

\subsubsection{Model Calibration}
Following \citep{whenDoes}, our work evaluates model calibration by formulating the generation process as the next token prediction task. Expected Calibration Error (ECE) and Maximum Calibration Error (MCE) of the models on the corpora tested are reported in Table \ref{table:evaluation1}.

  It is clearly demonstrated that the proposed methods, especially with sentence-level hard gate, lead to a calibrated student. The ECE and MCE scores of the base method are high as modern neural networks are found to be overconfident \citep{pmlr-v70-guo17a}. The amount of error is mitigated to some extent with the introduction of regularizers and KD methods. For instance, LS with uniform distribution lowers the ECE and MCE score to 6.43 and 9.98 on IWSLT14 corpus. The errors are reduced to around half of those of the base method. Nevertheless, the most noticeable gain in calibration is seen across the proposed methods. HKD-S achieves 1.27 in ECE and 3.22 in MCE, illustrating remarkable improvement in model calibration. The absolute improvement compared to the base method is approximately 11.7 score in ECE and 16 in MCE on IWSLT14 dataset. The proposed methods enhance the model calibration of a student by a large margin, with such gain observed across the corpora.
  \begin{figure}[t]
    \centering
    \includegraphics[width=0.9\columnwidth]{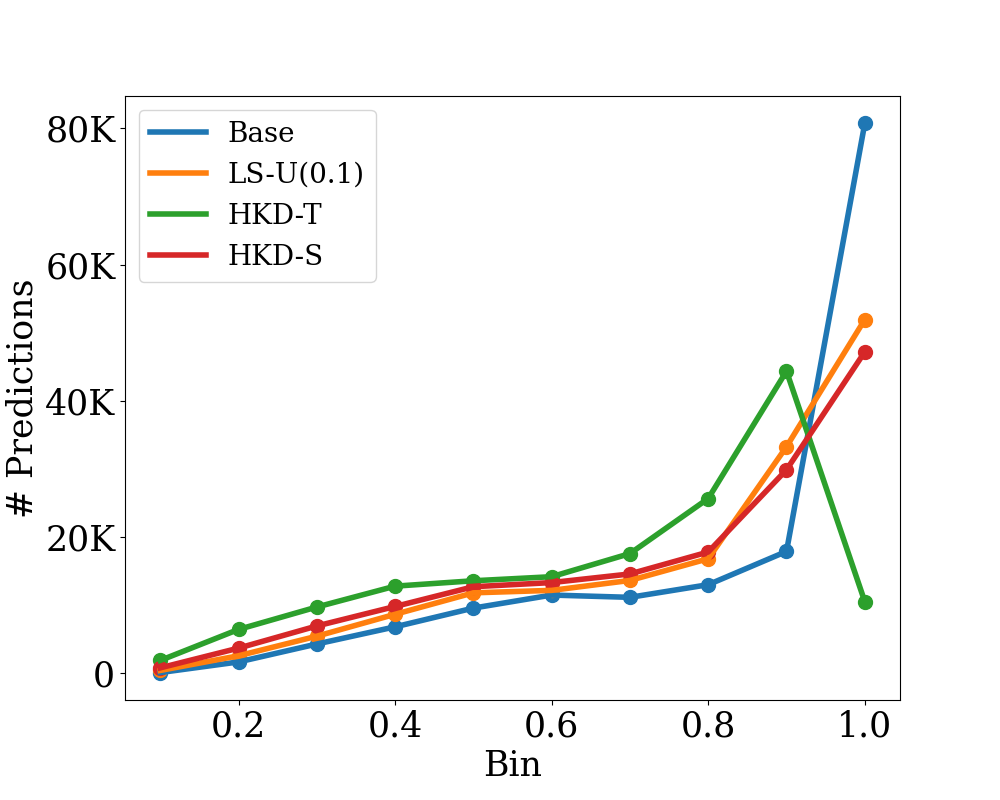}
    \caption{This figure shows the confidence histogram of the three approaches (Base, LS-Uniform, and HKD) on IWSLT14. The predictions are binned to 10 bins based on their confidence scores.}
    \label{fig:Histogram}
    \end{figure}
  Reliability diagrams in Figure \ref{fig:reliability} further support the claim. Despite the baselines improving model calibration, the methods tend to show either underconfident or overconfident predictions. LS-Uniform and PS-KD consistently make overconfident predictions, while SD suffers from underconfidence. On the other hand, the reliability diagrams of our work display calibrated results that mainly conform with the low ECE and MCE scores in Table \ref{table:evaluation1}.

Furthermore, our methods make predictions more evenly distributed as illustrated in Figure \ref{fig:Histogram}. HKD-T and HKD-S do not make predictions solely with low confidence; the number of predictions with confidence scores between 0.0 to 0.9 by our methods outnumbers those of the other methods, demonstrating an ability to make decision with \emph{diverse confidence}. 


\subsection{Ablation Study}
\begin{table}
  \centering
  \resizebox{0.75\columnwidth}{!}{
  \begin{tabular}{ccc}
  \hline
  \textbf{Method} &\textbf{BLEU}& \textbf{ECE}\\
  \hline
  Base & 35.96&12.98\\
  \hdashline
  $\phi$ trained with LS & 38.27&1.43\\
  $\phi$ trained with CE ($\tau$=1.0)& 37.79 & 5.87\\
  $\phi$ trained with CE ($\tau$=1.5) & 38.57 & 1.87\\
  \hline
  \end{tabular}}
  \caption{Ablation study on the proposed approach with different temperatures and knowledge. CE denotes training a model with the cross-entropy with hard targets.}
  \label{table:ablation}
  \end{table}
In order to further validate the proposed method, we conduct an ablation study with different teachers and varying temperature values, and the results are shown in Table \ref{table:ablation}. In a case where a teacher is trained with the cross-entropy with hard targets, both BLEU score and ECE score are enhanced compared to those of Base method. However, with a proper temperature control (enhanced calibration), both BLEU and ECE improve significantly. The BLUE score is higher than that of a model trained with LS-Uniform, and the ECE score is also competitive. This empirically validates that with a proper control of temperature, the proposed KD systems are compatible with a \emph{wide choice of teacher}.


\section{Conclusion}
In this study, we present hard gate knowledge distillation, a mechanism that switches supervision between knowledge and ground truth at either the sequence-level or the token-level. This originates from the novel view and role of a teacher in KD. The proposed method is simple yet effective in improving model generalization and calibration, achieving superior performances compared to those of the strong baselines.
\section*{Limitations}
As in previous KD methods, the proposed approaches utilize a teacher model, hence inevitably causing the computation cost to increase. In addition, as two forward passes are needed, one by a teacher and the other by a student, the training time is longer than non-KD training methods. Lastly, the proposed idea does not suit a natural language understanding task due to the introduction of the token-level and sentence-level gate.
\section*{Ethical Consideration}
The proposed idea is a student-teacher framework; hence, a teacher model can greatly affect a student model. If a teacher model is trained with a dataset with biased information or misinformation, the student is likely to learn such features while minimizing the knowledge gap. One can mitigate the concern to some extent if fact checking system or biased detection system is employed. This is not the fundamental solution to the problem that KD training faces, yet the level of danger is expected to be mitigated to some extent.
\section*{Acknowledgement}
Research on this paper was supported by Hong Kong Research Grants Council (Grant No. 16204920).

\bibliography{emnlp2022}

\bibliographystyle{acl_natbib}
\appendix
\section{Dataset \& Implementation Details}
\label{sec:Appen-data}
Multi30K dataset is the smallest in size among the corpora tested, 28K sentence pairs for training, 1K for validation, and 1K for testing. IWSLT15 EN-VI comprises 133K pairs in the training set, 1.5K in the validation, and 1.3K in the testing set. Lastly, IWSLT14 DE-EN corpus contains around 170K, 7K, and 7K pairs of sentences for training, validation, and testing dataset respectively. Words are processed into sequence of subword units with \texttt{subword-nmt} \citep{subword-nmt}\footnote{\url{https://github.com/rsennrich/subword-nmt}}.

The structure of models, both the student and the teacher, in all of the experiments follow transformer architecture \citep{transformer}. Specifically, both the encoder and the decoder are composed of 6 transformer layers with 4 attention heads, and the hidden dimension size is set to 512. The dropout probability is set to 0.3, and the maximum number of tokens within a batch is 4096. For the temperature in our experiments, we have tested \{0.8, 1.0, 1.5, 2.0, 2.5\}, and find that 1.0 works the best with a self-teacher trained with label smoothing. For a self-teacher trained with hard targets, 1.5 for temperature value illustrates the best performance, as the temperature smoothed the output of the teacher. We report random seeds used in our work for reproducibility, the seeds being \texttt{\{0000, 3333, 5555\}}. 
\end{document}